\def\year{2021}\relax
\documentclass[letterpaper]{article} 
\usepackage{aaai21}  
\usepackage{times}  
\usepackage{helvet} 
\usepackage{courier}  
\usepackage[hyphens]{url}  
\usepackage{graphicx} 
\usepackage{natbib}  
\usepackage{caption} 
\usepackage{amsfonts,amssymb}
\usepackage{amsmath}
\usepackage{booktabs}
\usepackage{comment}
\usepackage{microtype}

\title{Sketch and Customize: A Counterfactual Story Generator}

\author{
Changying Hao,\textsuperscript{\rm 1}\textsuperscript{,\rm 2}
  Liang Pang,\textsuperscript{\rm 1}\textsuperscript{,\rm 2}\thanks{Corresponding author.}
  Yanyan Lan,\textsuperscript{\rm 1}\textsuperscript{,\rm 2}
  Yan Wang,\textsuperscript{\rm 3}
  Jiafeng Guo,\textsuperscript{\rm 1}\textsuperscript{,\rm 2}
  Xueqi Cheng\textsuperscript{\rm 1}\textsuperscript{,\rm 2}
    \\
}
\affiliations{

    \textsuperscript{\rm 1}CAS Key Laboratory of Network Data Science and Technology, \\Institute of Computing Technology, Chinese Academy of Sciences\\
    \textsuperscript{\rm 2}University of Chinese Academy of Sciences\\
    \textsuperscript{\rm 3}Tencent AI Lab\\
    
    \{haochangying18s, pangliang, lanyanyan, guojiafeng,  cxq\}@ict.ac.cn, brandenwang@tencent.com
}

\begin{document}

\maketitle

\begin{abstract}


Recent text generation models are easy to generate relevant and fluent text for the given text, while lack of causal reasoning ability when we change some parts of the given text.
Counterfactual story rewriting is a recently proposed task to test the causal reasoning ability for text generation models, which requires a model to predict the corresponding story ending when the condition is modified to a counterfactual one. Previous works have shown that the traditional sequence-to-sequence model cannot well handle this problem, as it often captures some spurious correlations between the original and counterfactual endings, instead of the causal relations between conditions and endings.
To address this issue, we propose a sketch-and-customize generation model guided by the causality implicated in the conditions and endings. In the sketch stage, a skeleton is extracted by removing words which are conflict to the counterfactual condition, from the original ending. In the customize stage, a generation model is used to fill proper words in the skeleton under the guidance of the counterfactual condition. In this way, the obtained counterfactual ending is both relevant to the original ending and consistent with the counterfactual condition. Experimental results show that the proposed model generates much better endings, as compared with the traditional sequence-to-sequence model.

\end{abstract}


\section{Introduction}\label{sec:intro}

Conditional text generation has been a research hotspot in recent years, and various kinds of conditions such as context~\cite{sordoni2015neural,jaech2018low}, topic~\cite{xing2017topic,wang2018reinforced}, personality~\cite{li2016persona,yang2018investigating}, emotion~\cite{zhou2018emotional,fu2018style} have been used to guide the generation. Most of those conditional text generation tasks focus on how to generate good texts under certain conditions, while few work concentrates on how the consequence changes when the condition is modified. That is, the causal relationship between condition and its corresponding generation is not well studied in these tasks. 
To facilitate the study of causal reasoning in text generation, ~\citet{qin2019counterfactual} propose a novel counterfactual story generation task that cares more about how to revise an original story ending guided by a modified condition.

\begin{figure}
    \centering
    \includegraphics[width=1\linewidth]{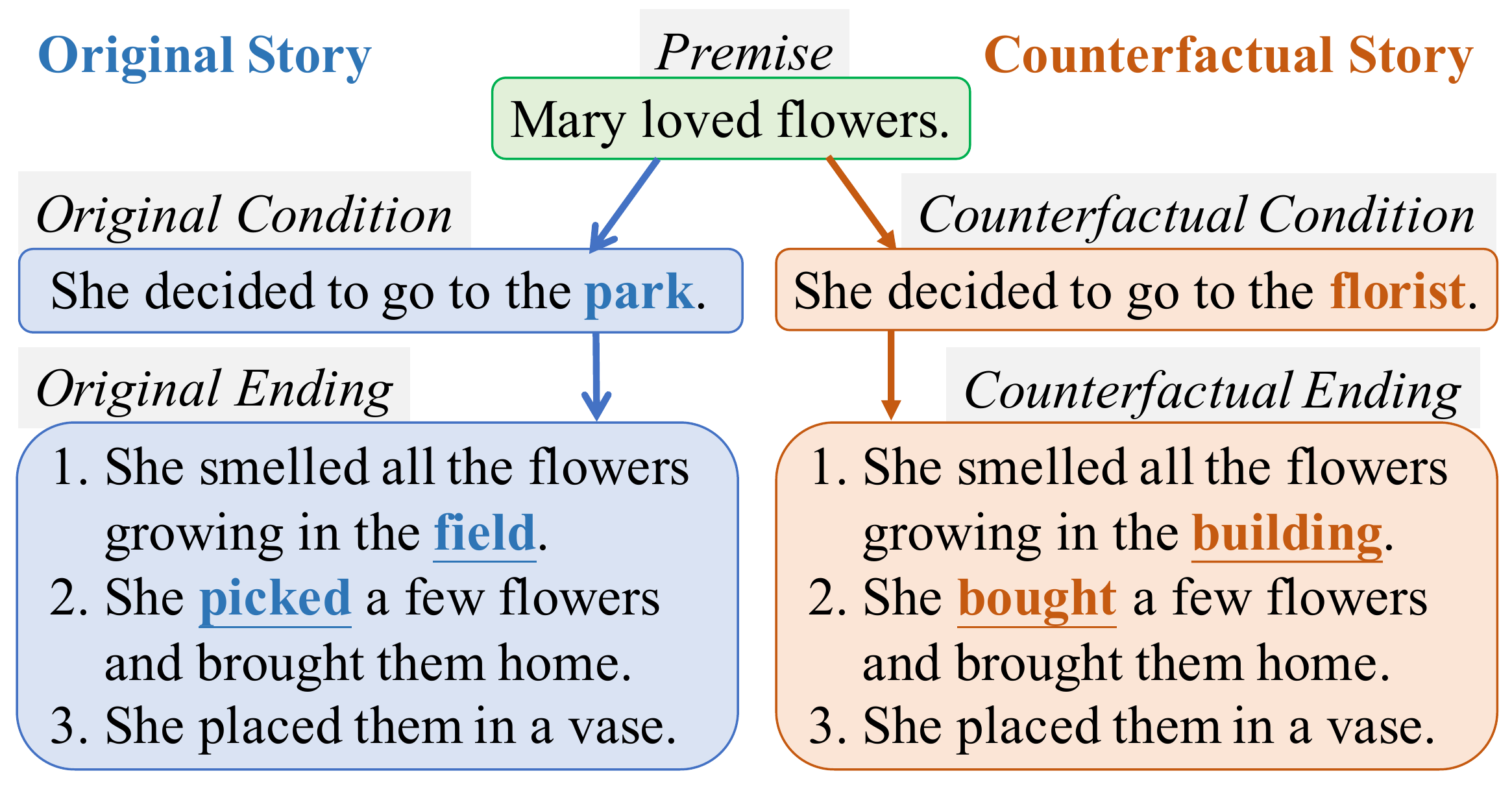}
    \caption{An example of an original story and counterfactual story pair.
    Black words in the two endings are background content, colored words are causal content.}
    \label{fig:storypair1}
\end{figure}
As shown in Figure~\ref{fig:storypair1}, in this dataset, given an entire original story (consisting of a one-sentence premise, a one-sentence condition, and a three-sentence ending) and an intervening counterfactual condition, some phrases in the original ending may conflict with the new condition. For example, in the new condition, ``she decided to go to the \textit{florist}'' instead of the \textit{park}. As a result, the place \textit{field} and the action \textit{picked} in the original ending should be replaced by some new words to eliminate the conflict. To rewrite a counterfactual ending compatible with the given counterfactual condition, an intelligent model should notice the condition changes and revise the conflict words in the original ending properly (\textit{field}$\rightarrow$\textit{building} and \textit{picked}$\rightarrow$\textit{bought} in Figure~\ref{fig:storypair1}).

\citet{qin2019counterfactual} use a standard sequence-to-sequence model to conduct the generation process. Specifically, this work concatenates the entire original story and the counterfactual condition as the source sequence and treats the counterfactual ending as the target sequence. 
Experimental results show that the model has difficulty in well capturing the causal relations between conditions and endings. For example, when the condition changes, the model often directly copies the original ending without knowing which part should be changed in the counterfactual ending. As a result, the generated counterfactual ending usually has some conflict with the counterfactual condition.

From the causal perspective, we can split both the original and counterfactual ending into two parts, the background content and the causal content. The background content is mostly related to the premise, which will not be influenced by the condition changes. 
While the causal content is caused by the difference between the two conditions, each causal content is relevant to the corresponding condition and conflict with the other condition.
Then, to generate counterfactual ending, a model needs to correctly detect the causal content and rewrite it to match the counterfactual condition. As can be seen from the example in Figure~\ref{fig:storypair1}, the condition modified the place from \textit{park} to \textit{florist}, the words in the causal content then change from \textit{field} to \textit{building}, and from \textit{picked} to \textit{bought}.

Taking the causality between the condition and ending into account, we propose a two-stage sketch-and-customize method that first sketches the story background and then customizes it with the counterfactual condition. In the sketch stage, according to the premise and two conditions,
the model detects the causal content from the original ending, through a sequence labelling process. Then the causal content will be masked and a skeleton that contains background content only should be preserved.
In the customize stage, the premise, counterfactual condition and the skeleton are provided as contexts to a generator to produce proper words to fill in the skeleton in a sequence-to-sequence way. Considering the fact that multiple rewriting ways are available for a given counterfactual condition, to improve the robustness of the generation model,
we apply several methods to augment the skeletons data, such as removing, replacing, and random shuffling some background words.

Experimental results show that our method significantly improves the consistency between the generated counterfactual ending and the counterfactual condition while keeping their relevance to the premise and the original ending. 
What's more, we make extensive comparisons of the counterfactual endings generated in different settings to show the effectiveness of the two-stage sketch-and-customize method.

\begin{figure}
    \centering
    \includegraphics[width=1\linewidth]{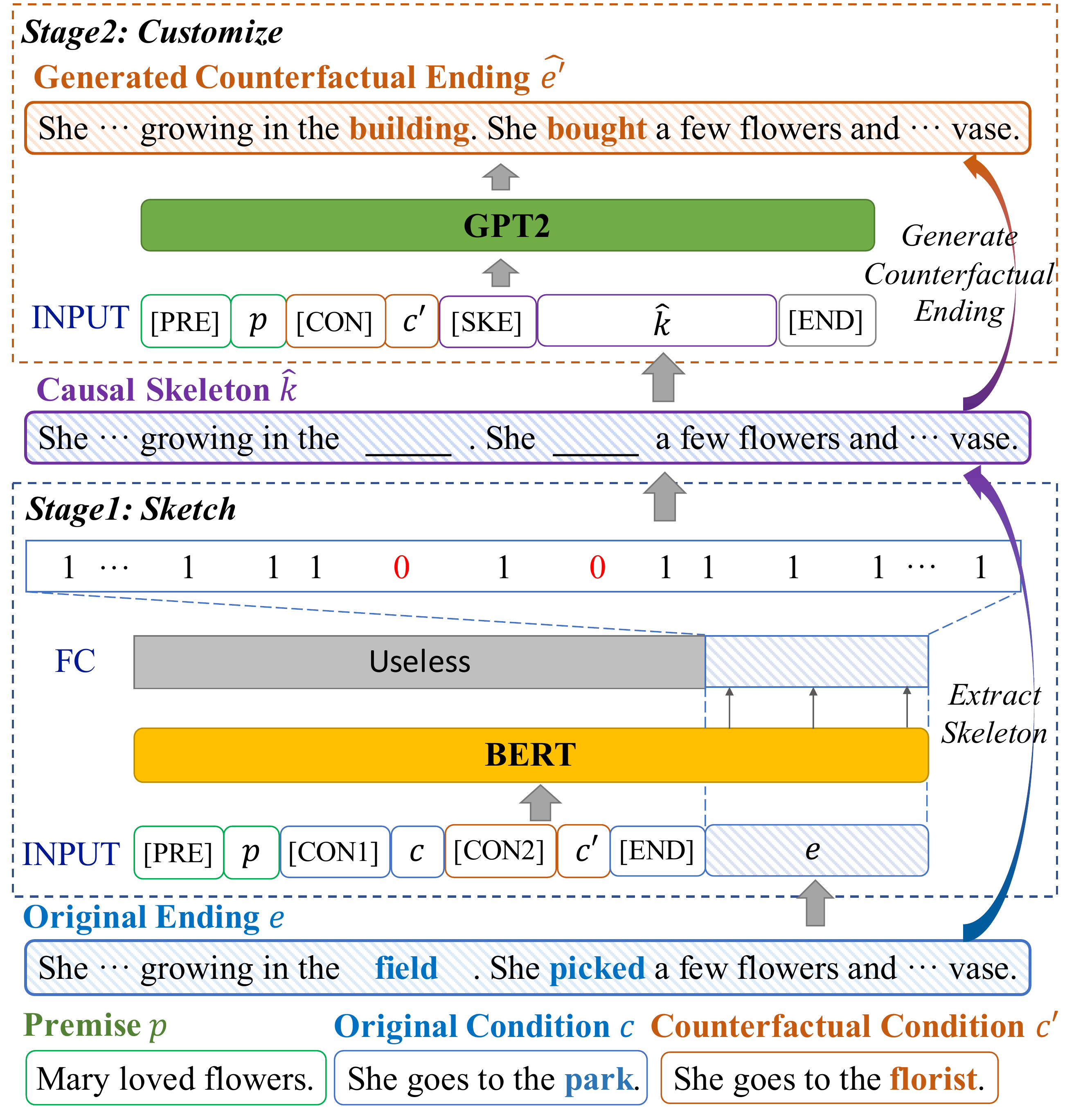}
    \caption{The structure of the two-stage sketch-and-customize model.}
    \label{fig:model}
\end{figure}

The main contributions of this paper include: 

1) Revisit the text generation task in a causal perspective, where the generated text is split into the background and causal parts, which is related to the premise and the changed condition respectively;

2) Propose a sketch and customize framework for improving the causal reasoning ability of the text generation models;

3) Conduct experiments on a counterfactual story rewriting task to verify the performance of our proposed framework.

\section{S\&C: Sketch-and-Customize Model}

In this section, we introduce our proposed sketch and customize model, which denoted as S\&C.
We first give the formulation of the counterfactual story generation and its notations. Then, we provide an overview of the model. Finally, the two stages are described in detail.

\subsection{Formulation and Notations}

A complete story is defined as a five-sentence text $\{p,c,e\}$, where the first sentence $p$ is the premise, the second sentence $c$ is the original condition, and the last three sentences constitute the original ending, abbreviated as $e$.
After given a counterfactual condition denoted as $c'$,
the task requires to revise the original ending $e$ into a counterfactual ending $e'$ which minimally modifies the original one and regains narrative coherency to the counterfactual condition.

\subsection{Model Overview}
As shown in Figure~\ref{fig:model}, our model contains two stages: sketch and customize. 
In the sketch stage, we introduce a casual skeleton extractor implemented as a sequence labelling model.
The causal skeleton can be obtained by replacing the causal content with blanks and only reserving the background content.
Specially, guided by the premise and two conditions, we use the pre-trained language understanding model BERT~\cite{devlin2019bert} to predict a causal skeleton by detecting whether each word in the original ending should be replaced with a blank or not. 
In the customize stage, GPT2~\cite{radford2019language} is used to generate the counterfactual ending by taking the premise, counterfactual condition and skeleton as inputs.

\subsection{Stage 1: Sketch}

As can be seen in Figure~\ref{fig:storypair1}, a reasonable story ending contains two parts, e.g. background content and causal content, which are related to the premise and condition respectively. 
Distinguishing the background content from the causal content is the target in the current sketch stage.
Guided by the causal skeleton, the generation model only needs to generate the causal content which is most related to the given condition.

In order to extract the proper causal skeleton, we model it as a sequence labelling task, where each word in the story ending has a label to indicate whether it belongs to background content or causal content.
With the help of the pretrained BERT, we can involve rich contextual information for each word.
Taking BERT as our base structure, we concatenate the premise $p$, the condition $c$, the counterfactual condition $c'$ and the story ending $e$ as the input context, and predict which words are causal words in the story ending. After that, the original BERT is fine-tuned by applying a classification task on each word, which is a standard sequence labelling setting.  

As shown in Figure~\ref{fig:model}, the sketch stage consists of three steps: the input formatting step, the representation step and the sequence labelling step. 
The input formatting step is used to transform the five-sentence story into an input sequence. To help BERT distinguish different parts of the story, we add some special tokens between them, denoted as $[\cdot]$. Thus, the input sequence can be represented as $S =$ \{[PRE] $p$ [CON1] $c$ [CON2] $c'$ [END] $e$\}.
For the representation step, we feed input sequence $S$ into BERT, and get the last layer representation $R=\{r_1, \cdots, r_{N_r}\} \in \mathbb{R}^{N_r \times E}$, where $r_i$ denotes the representation of $i$-th word in the input sequence, $N_r$ is the length of $S$, and $E$ is the dimension of the representation. 

In the sequence labelling step, only the representations of the ending part are fed into a fully connected layer, which projects each representation into a two-dimensional vector. Further, a softmax function is applied to get the probability distribution of the label of each word in the ending,
\begin{equation}
\begin{aligned}
    &p_1(l_{i}|S)=\mathrm{softmax}(Wr_i + b), \\
    &l_{i} = \left\{
       \begin{array}{lr}
          0, &  i \in e_{causal}\\
          1, &  i \notin e_{causal}\\
       \end{array},
	\right.
	i \in [N_e, N_r]\\
\end{aligned}
\end{equation}
where $N_e$ is the start word index of the story ending $e$,  $W$ and $b$ are parameters of the fully connected layer, and $e_{causal}$ is a set of causal word indexes in the story ending.
As the original ending and counterfactual ending are symmetrical, we can swap them and use counterfactual condition and ending as the original ones, thus we double the training examples.

The common question is how to obtain the ground truth causal skeleton.
Given an original ending and counterfactual ending pair $\{e,e'\}$, we can approximate the ground truth causal skeleton by using the Longest Common Subsequence (LCS) ~\cite{hirschberg1977algorithms} between them. 
The largest common parts of the two endings are seen as the background content $e_{bg}$, while the remaining parts in each ending are treated as causal content for the original ending ($e_{causal}$) and counterfactual ending ($e'_{causal}$) respectively.
In this way, we get the label for each word in the two endings and use them to train our model in the sketch stage.
By replacing the causal words in the ending with blanks and further merging the consecutive blanks into one blank, we get the approximating ground-truth causal skeleton, e.g. LCS skeleton.

\subsection{Stage 2: Customize}
In the customize stage, given the causal skeleton $k$, we are asked to complete the fill-in-the-blanks task to generate the story ending consistent with the counterfactual condition. 
We take the GPT2 as our base model, as shown in Figure~\ref{fig:model}, and organized the input sequence as follow.
The premise $p$, the counterfactual condition $c'$ and the causal skeleton $k$ are concatenated together and separated by special tokens, \{[PRE] $p$ [CON] $c'$ [SKE] $k$ [END]\}.
The concatenated contexts are then used as the input sequence of the GPT2 model, its target is the ground-truth counterfactual ending $e'$. 

Similar to the sketch stage, in the training process we construct two training instances for an original and counterfactual story pair, e.g. \{[PRE] $p$ [CON] $c'$ [SKE] $k$ [END]\} for generating counterfactual ending $e'$ and \{[PRE] $p$ [CON] $c$ [SKE] $k$ [END]\} for generating original ending $e$.
Please note that in this stage, we use [CON] to mark both original condition and counterfactual condition, use [END] to mark both original ending and counterfactual ending. In this way, the GPT2 model can learn to copy the background words from the skeleton and generate proper words in the blanks guided by a single kind of condition while not being disturbed by the other kind of condition.

The token [END] can be seen as a starting symbol of the decoding process, then GPT2 generates the ending token by token. The output word distribution probability is as follow,
\begin{equation}
    p_2(y_t|x,y_{<t}) ={\rm GPT2}(x,y_{<t}),
\end{equation}
where $y_t$ (can be $e_t$ or $e'_t$) is the $t$-th token after the [END] token, $y_{<t}$ represents the words between the token [END] and the $t$-th token, $x$ represents the words before the token [END], ${\rm GPT2}(z)$ is the function of getting the current step output distribution of the  GPT2 fed $z$ as its input.



\subsection{Training}
In the sketch stage, the LCS skeleton is used to approximate the ground truth causal skeleton. After that, the words in the causal content are labelled as 0, the words in the background content are labelled as 1, thus each word in the story ending has their ground-truth label.
Then, for the original story as an example, the premise $p$, the condition $c$ and $c'$, and the original ending $e$ are provided as the input sequence of BERT for the sequence labelling task.
In current story rewriting task, causal words are much less than background words, mainly because of the minimal changing requirement in the dataset annotation. 
Therefore, we adopt the weighted cross-entropy loss~\cite{xie2015holistically} and assign a larger weight for the causal words to overcome the data unbalance problem,
\begin{equation} \label{eq:seq_loss}
\begin{split}
    \mathcal{L}_{seq} = -\sum_{i=N_e}^{N_r}[\lambda &\log p_1(l_{i}=0|S)\\+(1-\lambda)&\log p_1(l_{i}=1|S)],\\
\end{split}
\end{equation}
where $N_r-N_e$ is the length of the original ending and $\lambda$ is the weight to control the loss of two labels.

In the customize stage, given the causal skeleton $k$, we train the generation model using the following loss,
\begin{equation}
    \mathcal{L}_{gen} = -\sum_{t=1}^{m}\log[p_2(e'_t|p,c',k,e'_{<t})],
\end{equation}
where $e'_t$ is the $t$-th word in the counterfactual ending, $e'_{<t}$ represents the words before the $t$-th word, $m$ is the length of the counterfactual ending.

\textbf{Causal Skeleton Augmentation:} 
Given a counterfactual condition, there are multiple ways to modify the existing original ending to fit the counterfactual condition. For example, in the validation and test set, three different endings are provided for each counterfactual condition. Therefore, the LCS skeleton may just reflect one possible rewriting intent. As a result, the generation models will be limited in learning various counterfactual endings, and hurt the generalization ability. To tackle this problem, we augment the current LCS skeleton with the following three alternative ways, 
1) randomly replacing 20\% of background words with blanks,
2) randomly replacing 20\% of background words with words sampled from the vocabulary,
3) randomly shuffling the order of background words in each skeleton.
After that, we merge the consecutive blanks into one blank and get augmented causal skeletons. During training in the customize stage, we combine the LCS skeleton and three kinds of augmented skeletons together to train the generation model. In this way, the model can be more compatible with noisy skeletons.

\subsection{Inference}
In the inference process, only the original story and the counterfactual condition are provided, we use the proposed two-stage sketch-and-customize model to generate the counterfactual ending.

In the sketch stage, the model predicts the label of each word in the original ending guided by the premise and the two conditions, the predicted label for each word in the original ending is calculated as follows,
\begin{equation}
    \hat{l_i} = \mathop{\arg\max}_{l_i\in \{0,1\}} \; p_1(l_i|S),
\end{equation}
the word is assigned as the label which has a max probability. Thus we get the causal words and replace them with blanks. Then we merge the consecutive blanks into one blank and get the predicted skeleton $\hat{k}$.

In the customize stage, using the premise $p$, the counterfactual condition $c'$ and the predicted skeleton $\hat{k}$, the GPT2 predicts the counterfactual ending token by token,
\begin{equation}
    \hat{e'_t} = \mathop{{\rm sample}}_{e'_t \in V} \; p_2(e'_t|p,c',\hat{k},e'_{<t}),
\end{equation}
where ${\rm sample}$ represents the top-k ~\cite{fan2018hierarchical} sampling method, $V$ is the vocabulary. When a sentence terminator is predicted, the decoding process is done and producing a generated counterfactual ending $\{\hat{e'_1},\hat{e'_2},\dots,\hat{e'_n}\}$ of length $n$.


\section{Experiments}

In this section, experimental settings are described and experimental results and detailed analyses are conducted on the proposed model.

\subsection{Dataset}
We use the large version of TimeTravel dataset proposed by ~\citet{qin2019counterfactual} as our dataset, it is built on top of the ROCStories~\cite{mostafazadeh2016corpus} corpus. The TimeTravel dataset contains 28,363 training original and counterfactual five-sentences story pairs. The development and test sets both have 1,871 original stories, each of the original stories in the development and test sets has a counterfactual condition and three rewritten counterfactual endings.

\subsection{Evaluation Metrics}


\subsubsection{Human Evaluation}\label{scetion:human_eval}

\citet{qin2019counterfactual} explore various of automatic metrics to evaluate the quality of the produced generations in their work, however, almost all of the scores of the automatic metrics correlate negatively with human scores in terms of the consistency to the counterfactual condition. 
Thus, we use human evaluation as the golden metric. 

We random sample 100 examples from the test dataset and ask four annotators to evaluate the quality of the counterfactual endings generated by different models, all of them are asked to give their scores in aspects of the relevance or similarity of the generated counterfactual ending to three components, i.e. the premise, the counterfactual condition, the original ending. The rules of human evaluation are finely described as follows,

\textbf{PRE} (Consistency and relevance to the premise):
If the generated ending keeps in mind details of the premise and does not in conflict with the premise anywhere, it should be given a score 3; If the ending is related to the premise, but there are few unimportant places where the word in the ending is in conflict with the premise, the score should be 2; If the ending completely violates the setting of the premise, the score should be 1.

\textbf{CF} (Consistency to the counterfactual condition):
If the ending is completely consistent with the counterfactual condition, the premise, the counterfactual condition and the generated ending can form a logically consistent story,
give score 3 to it; If the ending has a general consistency with the counterfactual condition, but there may be some minor conflicts between the ending and counterfactual conditions, it should be scored 2; If there are strong conflicts between the ending and the counterfactual condition, it should get score 1.

\textbf{PLOT} (Similarity to the plot of the original ending):
If the main body of the original ending is reserved in the generated ending, it should be scored 3; If the ending maintains a general relevance to the plot of the original ending, it should be scored 2; If the ending has nothing to do with the plot of the original ending, it should be scored 1.

\subsubsection{Automatic Evaluation}

The task requires minimal editing of the original ending to get a counterfactual one. ROUGE-L~\cite{lin2006information} is an appropriate metric which can reveal how much content of the original ending can be reserved in the generated ending by measuring the length of LCS between them.
We additionally use it to evaluate the similarity between generated endings and the original endings on the whole test set automatically.

\subsection{Implementation Details}

We first use the BasicTokenizer \footnote{https://github.com/huggingface/transformers\label{foot:huggingface}} to tokenize the training set and get a vocabulary of size 25044. The tokenized data are firstly used for LCS and the vocabulary is used for the skeletons augmentation. These basic tokenized sequences are further tokenized using BPE~\cite{gage1994new} method in both sketch stage and customize stage. The max sequence length is set to 300 with the padding token included. For the sketch model, we use the base uncased version of BERT\textsuperscript{\ref{foot:huggingface}}. The hidden size for the fully connected layer is 768. For the customize model, we use the medium version of GPT2\textsuperscript{\ref{foot:huggingface}}. We use Adam optimization for both models with initial learning rates set as 5e-5 and 1.5e-4 separately. The warmup strategy is applied with warmup-steps set to 2000. 
The batch sizes for the two stages are set to 8. 
We train the two-stage models for 5 and 10 epochs respectively and select the best models on the validation set.
During the inference in the customize stage, we use top-$k$ sampling with the temperature set to 0.7 and $k$ set to 40. The source code and all of the experiments can be found in \url{https://github.com/ying-A/SandC}.

\subsection{Compared Methods}
We make extensive comparisons for the counterfactual endings generated in different settings to verify the effectiveness of our model.
We compare the generated counterfactual endings using ~\citeauthor{qin2019counterfactual}'s method, two methods using random or LCS skeletons instead of generated skeletons, three kinds of S\&C methods, and ground-truth counterfactual endings. There will be seven generated counterfactual endings for each test instance, we do automatic evaluation on the whole test set. For human evaluation, we randomly sample 100 instances from the test set, the annotators are asked to given their evaluation scores in three aspects described in section~\ref{scetion:human_eval}.

\textbf{Seq2Seq-GPT}: The supervised Seq2Seq model implemented as ~\citeauthor{qin2019counterfactual} based on GPT2. It concatenates the original story, premise and counterfactual condition together as the input and treats the counterfactual ending as the output.

\textbf{Random\&C}: Randomly replace some words (the same number with the causal words in the original ending) with blanks in the original endings, then use the random skeletons for the customize stage. 

\textbf{LCS\&C}: Use the LCS skeletons for the customize stage.

\textbf{S\&C-$\lambda$}: The full version of our S\&C model, the $\lambda$ in the weighted cross-entropy loss (Eq~\ref{eq:seq_loss}) is set to 0.5 or 0.8.


\textbf{S\&C-w/o-Aug}: It does not use the augmented skeletons data when training in customize stage. The loss weight for the causal words is set to 0.8.

\textbf{Human} One of the three ground-truth counterfactual endings edited by human.

\subsection{Main Results}

\begin{table}
    \centering
    \small
    \begin{tabular}{lllllll}
        \midrule[0.75pt]
            \textbf{$\lambda$}  & \textbf{CP} & \textbf{CR} & \textbf{CF1} & \textbf{BP} & \textbf{BR} & \textbf{BF1}\\
        \midrule[0.75pt]
        
        0.2 &    0.84   &  0.13     & 0.22 &0.78 &0.99 & 0.87 \\
        \hline
        
        0.5 &  0.60    &  0.33    & 0.43& 0.81 &0.93 & 0.87\\
        \hline
        0.8&0.41&0.68&0.51&0.87&0.69&0.77\\
        
        \midrule[0.75pt]
    \end{tabular}
    
    \caption{
        The prediction scores of the sketch stage using the LCS as the golden skeleton, more details are stated in section 2.3,
        where $\lambda$ is the parameter in the loss function, C means causal words, B means background words, P,R,F1 are precision, recall, F1-score respectively. For examples, CP means causal words precision, BF1 means background words F1.
    }
    \label{tab:first_eval}
\end{table}


For the first stage, we train the sketch model under three $\lambda$ settings of the loss function, which are 0.2,0.5 and 0.8 respectively. The prediction scores are given in Table~\ref{tab:first_eval}. 
We can see that the Precision-Recall shows a trade-off for both causal words and background words. What we emphasize most is the Recall for the causal words (CR), as we should make sure that the causal words in the original endings are mostly removed, so that the input skeleton to the generation model will include few causal words which is inconsistent with the counterfactual condition. When $\lambda$ is set to 0.8, we get the best CR and CF1. Thus we simply compare the models with $\lambda$ setting to 0.5 and 0.8 in the following customize stage.

\begin{table}
    \centering
    \small
    \begin{tabular}{lllll}
    \midrule[0.75pt]
          & \textbf{PRE} & \textbf{CF} & \textbf{PLOT} & \textbf{Avg.}\\
    \midrule[0.75pt]
    Seq2Seq-GPT &    2.558   &  1.985$^\downarrow$     & 2.170 &2.238 \\
    \hline
    Random\&C &  2.572     &  1.905$^\downarrow$     & 2.132& 2.203\\
    \hline
    LCS\&C &  2.542     &  2.083    & 2.145& 2.257\\
    \hline
    S\&C-0.5 &   \textbf{2.650}   &  1.668$^\downarrow$     &  \textbf{2.425}$^\uparrow$   &2.248\\
    \hline
    S\&C-0.8 &  2.590    &  \textbf{2.130}   & 2.120 &$\textbf{2.280}$\\
    \hline
    S\&C-w/o-Aug &  2.458$^\downarrow$  &2.030     &  1.845$^\downarrow$ &2.111\\
    \hline
    Human &   2.610    &   2.217    &  2.252$^\uparrow$  &2.360\\
    \midrule[0.75pt]
    \end{tabular}%
    \caption{
    Human evaluation scores on generated counterfactual endings for different methods. 
    Avg. is calculated by averaging the PRE, CF and PLOT score. 
    The top-performing model except for ``Human'' is bolded.
    Significant tests (t-test) are performed based on S\&C-0.8, the method whose score significantly lower than S\&C-0.8 is marked with $^\downarrow$, higher than S\&C-0.8 is marked with $^\uparrow$($p$-value$<0.01$).
    }
    \label{tab:human_eval}
\end{table}

\begin{table*}
    \centering
    \small
        \begin{tabular}{cllllllll}
        \toprule[0.75pt]
        Model & Seq2Seq-GPT & Random\&C & LCS\&C & S\&C-0.5 & S\&C-0.8 & S\&C-w/o-Aug & Human \\
        \midrule
        ROUGE-L & 0.899 & 0.806 & 0.79  & 0.921 & 0.644 & 0.597 & 0.791 \\
        \bottomrule[0.75pt]
        \end{tabular}%
    \caption{ROUGE-L between generated counterfactual endings and original endings for different methods.}
    \label{tab:my_label}
\end{table*}
To verify the superiority of our proposed two-stage sketch-and-customize model, we compare our model with the recent Seq2Seq-GPT baseline model in the three aspects comprehensively.
As can be seen from the human evaluation results depicted in Table~\ref{tab:human_eval}, our S\&C-0.8 model get the highest CF score among all the methods, it is 0.145 higher than Seq2Seq-GPT, which is a significant improvement. It reveals that our sketch-and-customize model can generate more consistent counterfactual endings to the counterfactual condition and avoid simply copying the original ending as the generated ending. Meanwhile, it also outperforms Seq2Seq-GPT on PRE, which demonstrates that our method can generate consistent and relevant endings to the premise. 
The PLOT is used to evaluate the word-level similarity between the generated ending to the original ending, our S\&C-0.8 model is evaluated less similar with the original ending compared to Seq2Seq-GPT baseline. 
The main reason is that Seq2Seq-GPT often applies a copy strategy from original ending, that yields a nearly perfect similarity score, while our model needs detecting the causal words and properly revising them, thus the generated endings by Seq2Seq-GPT can have more overlap to the original endings than by S\&C-0.8 model. 
To assess model in a comprehensive way, we average PRE, CF and PLOT to get an overall score, denoted as Avg. Our S\&C-0.8 method gets the highest score on the average of three aspects.


\subsection{More Analysis}


Additionally, comparative experiments are conducted to verify the effectiveness of each component and hyper-parameters in S\&C model.

\textbf{Influence of the loss weight for causal words.}
Comparing the performances of S\&C-0.5 and S\&C-0.8 where the loss weight $\lambda$ are set to 0.5 and 0.8 respectively, we find that S\&C-0.8 outperforms S\&C-0.5 on the CF metric significantly.
It reveals that hyper-parameter $\lambda$ is important for the task, and the reasons can be drawn into two aspects.
Firstly, the label imbalance problem occurs in the training of the sketch stage, where the ratio of causal words and background words are 1:4, due to the minimal changing requirement in this task. To balance the labels, multiplying four times weight to the loss of causal words leads to performance improvement.
Secondly, higher loss weight causes skeletons with more blanks. It improves the recall of the causal words and reduces the conflicts between the predicted skeleton and the counterfactual condition correspondingly, leaving more spaces for generation model to generate relevant content.  

\textbf{Influence of the causal skeleton.}
To examine the effectiveness of the causal skeleton, we remove the sketch stage in Random\&C and LCS\&C, instead, they get the causal skeleton by randomly replacing words with blanks in the original ending or directly using the LCS between two endings respectively.
As we can see in Table~\ref{tab:human_eval}, Random\&C performs poorly on CF compared to S\&C-0.8. Randomly removing words in original endings to obtain causal skeletons, there will still remain some causal words conflicting with the counterfactual condition. It shows that the causal words are important for generating counterfactual condition consistent endings. Moreover, the sketch stage in S\&C has the ability to remove causal words effectively.

Besides, we find that S\&C-0.8 behaves similarly to the LCS\&C on CF and PRE, or even slightly better. This is because the LCS skeleton is acquired by exactly removing causal words in the original ending, which is too strict for the generation model, and thus fails to fill the correct words in the blanks. In contrast, the skeleton got by S\&C-0.8 has a few more blanks, which preserves enough ``freedom'' for the generation process.


\textbf{Influence of skeletons augmentation.}
Comparing the human scores for S\&C-0.8 and S\&C-w/o-Aug, we find that S\&C-0.8 achieves better scores in all of these three aspects, which demonstrates the efficiency of using various kinds of causal skeletons to train the generation model. Using only the LCS skeletons to train the customize stage leads to overfitting and limits model generation capabilities.


\textbf{Relevance to the original endings.}
ROUGE-L between generated counterfactual endings and original endings for different methods are shown in Table~\ref{tab:my_label}. S\&C-0.8 model gets a score of 0.644 which is 0.147 lower than human but is acceptable. That means our generated counterfactual endings still have a large overlap in LCS between original endings. It respects the minimal editing requirement but still has an improving space.

\subsection{Case Study}
\begin{figure*}
    \centering
    \includegraphics[width=1\linewidth]{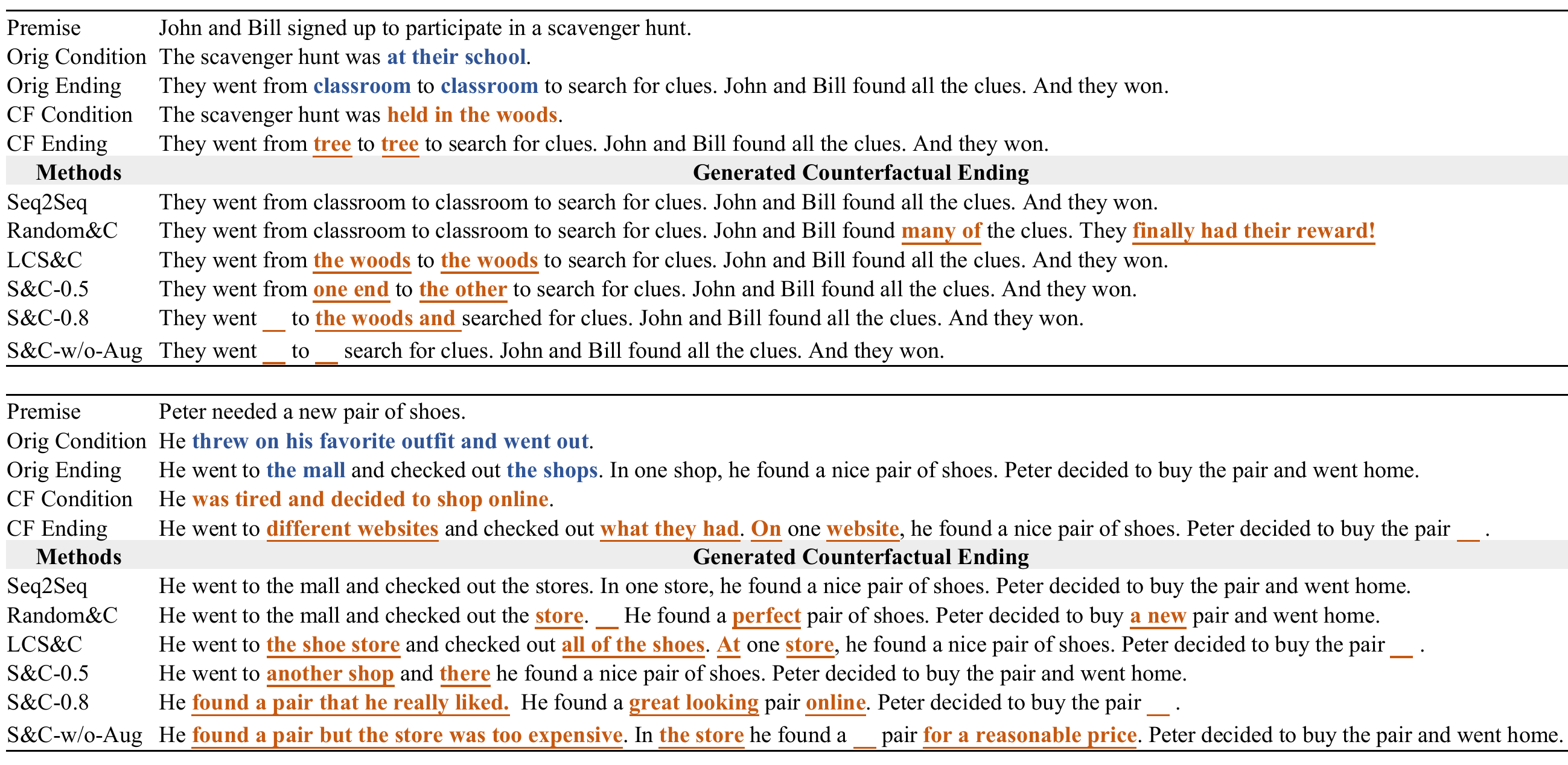}
    \caption{Two cases of counterfactual endings generated by different methods. Black words together with orange blanks represent the predicted skeletons by different methods, orange words represent the generated causal words.}
    \label{fig:case}
\end{figure*}
Figure~\ref{fig:case} presents two examples of generated counterfactual endings by different methods. The black words together with orange blanks are the skeletons predicted by different methods. The orange words are generated causal words in the customize stage.
In the two conditions of the first example, the place of the game changes from \textit{school} to \textit{woods}, the causal word then changes from \textit{classroom} to \textit{tree} in the ground-truth endings. As we can see, Seq2Seq-GPT directly copies the original ending as the counterfactual ending, which conflicts with the counterfactual condition. The Random\&C method also generates conflict ending using the randomly extracted skeleton. The LSC\&C method generates \textit{the woods} in the blanks, which is related to the counterfactual condition, but \textit{from the woods to the woods} is ungrammatical. 
As for S\&C methods, all of them successfully masks the inconsistent causal word \textit{classroom} and S\&C-0.8 method generates a consistent story ending. It reveals that the proposed method properly captures the fine-grained relation between the condition and the ending. 
In the second example, Peter decided to \textit{shop online} instead of in \textit{mall} in the counterfactual condition. S\&C-0.8 method successfully rewrites the ending to a consistent one. In contrast, even the LCS\&C method uses the LCS skeleton, it still generates conflict words in the blanks. Maybe it is because a too strict constraint affects the generation model's free exploration, and thus fail to fill proper words into the skeleton.

\section{Discussion}

Considering the way for a human to predict the consequence changes when the condition changes, we will use the causal structure as the prior knowledge, first detect the causal part and then modify it to a consistent one. Thus we propose a sketch and customize framework which identify the causal relations between the conditional text and the generated text in the conditional text generation task. Our findings can be generalized to other conditional text generation task. That is to say, the conditional text only affects the words which have the causal relationship with it, while does not influence the words which relate to the premise. However, the sequence to sequence framework does not distinguish those words and generates them as a whole. 

The counterfactual story generation dataset TimeTravel gives a good way to explore the causal relation, because there are multiple conditional texts for a same premise that yield different generated text (story ending). Experiments show that if we model the causal dependence clearly rather than simply fit the data as the traditional sequence to sequence framework, the model can generate better texts. We can design more comprehensive causal models or utilize the common-sense knowledge to guide the generation in the future.


\section{Related Work}

\textbf{GPT2 for text generation} Finetuning the pretrained GPT2 on specific generation tasks for a better performance turns to be a new trend in recent studies. 
GPT2 is a large transformer-based language model which is trained on a sufficiently large corpus. 
Given all of the previous words, its training objective is to predict the next word. 
Due to its superior generation ability, it has been applied in variety of text generation tasks, such as dialogue response generation~\cite{zhang2020dialogpt}, news generation~\cite{zellers2019defending}, and traditional story generation~\cite{guan2020knowledge,mao2019improving}. These works use GPT2 to generate sequences in a direct left-to-right way. 
Different from these works, we explore the ability of language modelling for GPT2 in a Fill-in-the-Blanks task. The well generated counterfactual endings in our paper indicate that GPT2 can generate proper words in blanks.\\
\textbf{Multi-step text generation} The proposed two-stage sketch-and-customize model is inspired by the recent success of the multi-step generation models. For example, in the dialogue generation task, \citet{cai2019skeleton,cai2019retrieval} first retrieve an initial similar response to the query, then delete conflict words from it and rewrite it to the final response according to the query; \citet{song2020generate} first generate a response to the query and then delete conflict words violating the personality and rewrite it to a personality-consistent response. In the story generation task, \citet{xu2018skeleton, yao2019plan} first generate key phrases or storyline of the story, then expand them to a complete sentence. However, the skeletons or keywords in their works are all got from heuristic methods. 
In our work, the skeletons can be predicted from a learnable model, benefiting from the distinctions between original and counterfactual conditions and endings, which can be treated as a strong supervised signal.

\section{Conclusion}
In this paper, we propose a novel two-stage sketch-and-customize model for counterfactual story rewriting task. Different from the traditional sequence-to-sequence model, S\&C explores the causal structure between the premise, condition and ending, and separate the story ending to two parts, the background content and the causal content. 
In the sketch stage, a skeleton is extracted to only contain the background content, by removing the causal content from the original ending. Then in the customize stage, a generation model is used to fill proper causal words in the skeleton under the guidance of the counterfactual condition.
Experimental results show that our model significantly improves the consistency between the generated counterfactual ending and the counterfactual condition while keeping their relevance to the premise and the original ending.

In this process, we have found that skeleton augmentation strategies such as removing, replacing, and random shuffling some background words are crucial to improve the robustness and generalization of our model. That is because there are usually multiple rewriting ways for a given counterfactual condition, and a strict skeleton with respect only one rewriting intent is limited to capture the implicated causality. In future, we plan to further study the story rewriting task from the perspective of causality, such as exploring confounders in the causal structure to reduce bias, and designing new causal models to generate more consistent and diverse counterfactual endings.

\section * { Acknowledgments}
This work was supported by the National Natural Science Foundation of China (NSFC) under Grants No.61906180 and 61773362, the Tencent AI Lab Rhino-Bird Focused Research Program (No.JR202033), the Lenovo-CAS Joint Lab Youth Scientist Project.

\bibliography{aaai2021}

\end{document}